\documentclass[11pt]{article}

\usepackage[preprint]{acl}

\usepackage{times}
\usepackage{latexsym}

\usepackage[T1]{fontenc}

\usepackage[utf8]{inputenc}

\usepackage{microtype}

\usepackage{inconsolata}

\usepackage{graphicx}
\usepackage{pgfplots}
\usepgfplotslibrary{groupplots}
\pgfplotsset{compat=1.18}

\usepackage{amsmath}
\usepackage{amsfonts}
\DeclareMathOperator*{\argmin}{arg\,min}
\DeclareMathOperator*{\argmax}{arg\,max}

\usepackage{booktabs}
\usepackage{multicol}
\usepackage{multirow}
\usepackage{tabularx}

\usepackage{needspace}
\usepackage{enumitem}
\usepackage[ruled,vlined,linesnumbered]{algorithm2e}
\usepackage{setspace}
\usepackage{xspace}
\usepackage{xcolor}
\usepackage[most]{tcolorbox}
\newcommand{\ourmethod}{\textsc{AlignEvoSkill}\xspace}
\definecolor{softred}{RGB}{250,100,100}
\definecolor{softgreen}{RGB}{56,118,29}
\definecolor{softblue}{RGB}{100,150,200}
\definecolor{softgray}{RGB}{150,150,150}

\newtcolorbox[auto counter, number within=section]{prompt}[2][]{%
  colback=white, 
  colframe=softblue!150, 
  width=\textwidth, 
  arc=3mm, 
  boxrule=0.8mm, 
  title=\normalsize #2, 
  breakable, 
  fonttitle=\small, 
  fontupper=\footnotesize, 
  #1 
}

\title{\ourmethod: Towards Knowledge-Aware and Task-Aligned \\ Agent Skill Evolution}

\author{
    Dingzirui Wang$^1$, Xuanliang Zhang$^1$, Keyan Xu$^1$ \\
    \bf{Qingfu Zhu$^1$, Wanxing Che$^1$, Yang Deng$^2$} \\
    $^1$Harbin Institue of Technology \quad $^2$Singapore Management University \\
    \{dzrwang, xuanliangzhang, kyxu, qfzhu, car\}@ir.hit.edu.cn \\
    ydeng@smu.edu.sg
}

\begin{document}
    \maketitle
    \begin{abstract}
        Reusable skills play a key role in improving LLM-based agents, but existing skill-evolution methods often fail to ensure that evolved skills both cover the knowledge required by the task and remain aligned with the target task. 
        As a result, evolved skills could be incomplete or irrelevant. 
        To address this limitation, we propose \ourmethod, a skill-evolution framework that jointly models knowledge coverage and task alignment. 
        Given failed task trajectories, \ourmethod first identifies task-relevant knowledge tags, retrieves complementary prior skills, and adapts them into candidate skills that address missing knowledge. 
        It then selects high-quality candidates using a joint filtering criterion based on knowledge-coverage and task-alignment scores. 
        Experiments on $3$ benchmarks with $4$ LLM backbones show a $34.7\%$ relative gain of \ourmethod over the non-evolution baseline and achieves a new SOTA in skill evolution with lower cost.
    \end{abstract}

    \section{Introduction}
        Agent skills have recently been introduced as reusable units of procedural knowledge that enhance the long-term capabilities of large language model (LLM) agents~\cite{xu2026agentskillslargelanguage}. 
By maintaining skills in external libraries, agents can retrieve and reuse prior experience for new tasks. 
However, manually annotating such skills is expensive, which has motivated recent studies on skill evolution, where agents automatically refine skill libraries from past experiences by generating reusable skills, leveraging both the trajectory and the existing skill library \cite{gao2026a}. 
Existing methods can be broadly grouped into two categories: skill-driven methods, which organize and expand existing skills~\cite{liu2024skillact,chen2026cuaskilldevelopskillscomputer,yang2026autoskillexperiencedrivenlifelonglearning}, and trajectory-driven methods, which distill execution traces and failures into reusable skills~\cite{wang2024voyager,chen-etal-2024-automanual,wang2025inducing,ni2026trace2skilldistilltrajectorylocallessons}.

\begin{figure}[t]
    \centering
    \small
    \includegraphics[width=\linewidth]{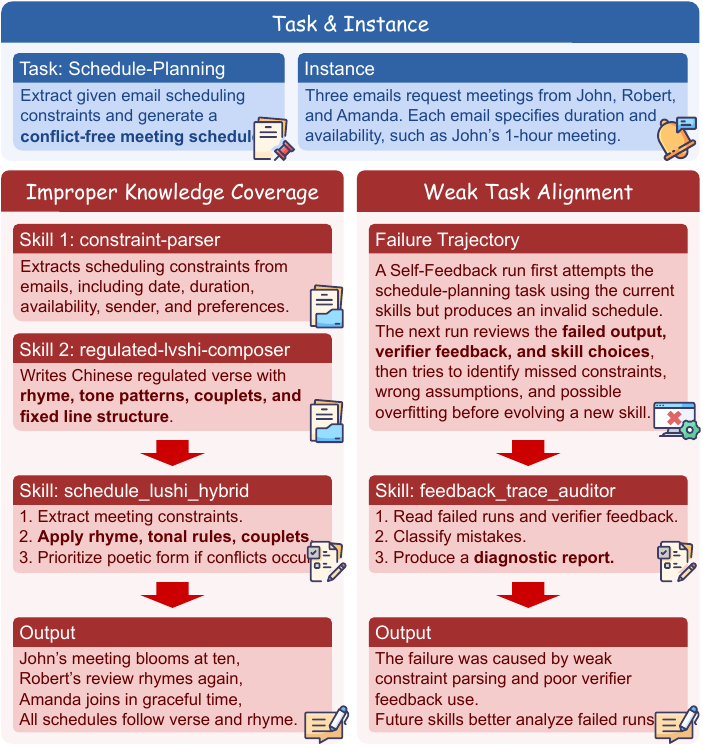}
    \caption{
        Illustration of two failure cases in current agent skill-evolving methods.
        The cases are sampled from SkillLearnBench and abridged for clarity.
    }
    \label{fig:motivation}
    \vspace{-1em}
\end{figure}

Despite recent progress, current skill-evolving methods still face two key limitations, as shown in Figure~\ref{fig:motivation}.
\textit{(i) Improper knowledge coverage:}
When evolving a new skill from existing skills, most methods either rely on a single skill or simply concatenate multiple skills. 
The former may overlook task-relevant knowledge scattered across different skills, while the latter may introduce irrelevant or conflicting information. 
Consequently, the evolved skill can be incomplete or noisy, limiting its usefulness for the target task~\cite{liu2026agenticskillsworkwild}.
\textit{(ii) Weak task alignment:}
When evolving skills from trajectories, existing methods mainly summarize, compress, or repair the observed execution process. 
However, focusing too heavily on the trajectory may cause the evolved skill to reflect the execution history rather than the actual task requirements. 
As a result, the skill may appear reasonable with respect to the trajectory but still fail to guide the agent toward solving the target task~\cite{zhong2026skilllearnbenchbenchmarkingcontinuallearning}.

To overcome these limitations, we propose knowledge-aware and task-alignment agent skill evolution (\ourmethod), a skill-evolving framework that improves evolved skills through knowledge-tag-guided generation and likelihood-based task alignment. 
\ourmethod is built on two key designs.
\textit{(i) Knowledge-tag-guided skill evolution:}
We associate agent skills with knowledge tags and use these tags to identify task-relevant prior knowledge. 
Given a target task and its trajectory, \ourmethod first extracts the required knowledge tags and then evolves new skills from existing skills that share these tags. 
This design enables effective transfer from relevant skills while reducing irrelevant or noisy information.
\textit{(ii) Likelihood-based task alignment:}
We further evaluate whether an evolved skill is aligned with the target task by estimating the likelihood of generating the skill conditioned on the task description. 
\ourmethod retains only highly aligned candidate skills, ensuring that the evolved skills provide task-specific and effective guidance.

We evaluate \ourmethod on $4$ widely used LLMs and $3$ skill benchmarks. 
Experimental results show that \ourmethod leads to a $34.7\%$ relative gain over the non-evolution baseline and establishes a new state-of-the-art (SOTA) result in skill evolution at a lower cost.
Further analysis shows that \ourmethod improves the knowledge-tag coverage score for task-relevant knowledge by $9.9\%$ and increases the task-alignment likelihood by $8.9\%$ over the strongest baseline. 
These results demonstrate that \ourmethod effectively alleviates improper knowledge coverage and weak task alignment.

Our contributions are summarized as follows:
\begin{itemize}[leftmargin=*,nosep]
    \item We identify two key limitations of current agent skill-evolving methods, namely improper knowledge coverage and weak task alignment, and propose \ourmethod to address such limitations through knowledge-tag-guided generation and likelihood-based task alignment.
    \item We conduct comprehensive experiments on $4$ LLMs and $3$ agent skill benchmarks. Results show that \ourmethod leads to a $34.7\%$ relative gain over the non-evolution baseline and achieves a new SOTA result with lower cost.
    \item Compared with existing skill-evolving baselines, further analysis shows that \ourmethod improves knowledge coverage and task alignment by $9.9\%$ and $8.9\%$, respectively, demonstrating its ability to mitigate the above limitations.
\end{itemize}

    \section{Related Work}
        Skill evolution aims to continually improve external skill libraries by revising, composing, specializing, and extending reusable procedural knowledge~\cite{xu2026agentskillslargelanguage}. 
Existing studies can be broadly divided into trajectory-driven and skill-driven approaches. 
Trajectory-driven approaches treat task trajectories as the primary source for skill acquisition. 
Representative systems such as ExpeL~\cite{zhao-etal-2024-expel} and Voyager~\cite{wang2024voyager} convert successful executions, failures, environmental feedback, and self-verification into reusable lessons or executable skills. 
Later studies further extract structured knowledge from interaction traces: AutoManual~\cite{chen-etal-2024-automanual} transforms interactive experiences into task manuals, Agent~Skill~Induction~\cite{wang2025inducing} derives verified programmatic skills from web trajectories, and Trace2Skill~\cite{ni2026trace2skilldistilltrajectorylocallessons} distills trajectory-local lessons into transferable skills. 
These methods ground skill construction in concrete executions, but their dependence on observed trajectories may cause evolved skills to capture trace-specific patterns rather than the knowledge actually required for the target task.

Skill-driven approaches focus on refining skills within the library. 
SkillAct~\cite{liu2024skillact} studies skill abstractions to improve interactive agents, while CUA-Skill~\cite{chen2026cuaskilldevelopskillscomputer} builds a parameterized skill base for computer-use agents. 
Recent systems such as AutoSkill~\cite{yang2026autoskillexperiencedrivenlifelonglearning} and SkillX~\cite{wang2026skillx} further maintain explicit skill repositories through continual extraction, refinement, hierarchical organization, and expansion from agent experiences. 
Although these methods support long-term skill reuse, they still face a coverage–alignment trade-off: adapting a single retrieved skill may overlook task-relevant knowledge distributed across multiple skills, while directly combining multiple skills may introduce irrelevant or conflicting information.

In contrast, \ourmethod jointly addresses improper knowledge coverage and weak task alignment. 
It uses knowledge tags to identify task-relevant requirements and retrieve complementary prior skills, thereby transferring missing knowledge while reducing noisy information. 
It further applies a task-conditioned likelihood criterion to filter candidate skills, ensuring that retained skills are both knowledge-complete and well aligned with the target task, thus effectively guiding future tasks.

    \section{Task Formulation}
        We formulate \emph{agent skill evolution} as improving an LLM-based agent whose parameters remain fixed, by updating only its external library of reusable skills. 
Let $\pi_\theta$ denote the base agent with frozen parameters $\theta$, and let $\mathcal{L}$ denote its skill library. 
Given a task $x$, the agent retrieves and applies relevant skills from $\mathcal{L}$ to generate an execution trajectory:
\[
\tau \sim \pi_\theta(\cdot \mid x, \mathcal{L}),
\]
where a task evaluator $q(x,\tau) \in \{0,1\}$ returns $1$ if the task is successfully completed and $0$ otherwise.

Starting from an initial skill library $\mathcal{L}_0$, agent skill evolution iteratively updates the skill library using experiences collected from evolution tasks. 
At each evolution epoch $r$, the agent collects a set of trajectories together with their feedback, denoted by $\mathcal{E}_r$. 
An evolution procedure $\mathcal{U}$ then uses these experiences, together with the current library, to produce an updated skill library:
\[
\mathcal{L}_{r+1} = \mathcal{U}(\mathcal{L}_r, \mathcal{E}_r).
\]
Since the base agent is frozen, any performance improvement must come from better skill acquisition, revision, or reuse within $\mathcal{L}_{r+1}$.

The goal is to obtain a final skill library $\mathcal{L}_R$ that improves the frozen agent's generalization performance on unseen test tasks:
\[
J(\mathcal{L}_R)
=
\mathbb{E}_{x \sim \mathcal{D}_{\mathrm{test}}}
\mathbb{E}_{\tau \sim \pi_\theta(\cdot \mid x, \mathcal{L}_R)}
[q(x,\tau)].
\]
During evolution, only trajectories and feedback from evolution tasks are available; no test-task trajectories or feedback can be used.

    \section{\ourmethod: Knowledge-Aware and Task-Aligned Skill Evolution}\label{sec:method}
        \begin{figure*}[t]
    \centering
    \small
    \includegraphics[width=\linewidth]{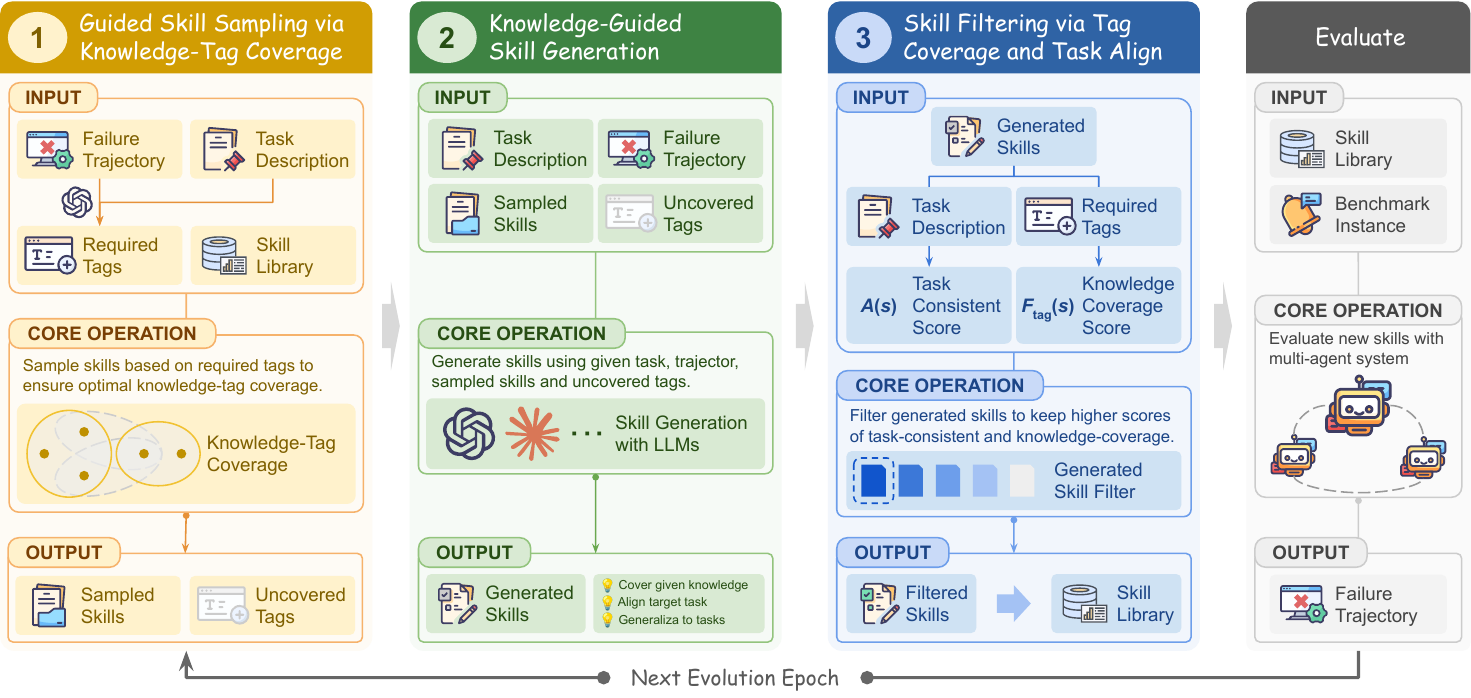}
    \caption{
        Illustration of \ourmethod, which instantiates the skill evolution operator $\mathcal{U}$ in three steps.
        \textit{(i) Knowledge-tag-guided source skill selection}: infer required tags and retrieve covering prior skills.
        \textit{(ii) Knowledge-tag-guided skill evolution}: generate candidates from selected skills, failed trajectories, and uncovered tags.
        \textit{(iii) Skill filtering by knowledge coverage and task alignment}: retain high-scoring candidates and update the library.
    }
    \label{fig:method}
    \vspace{-1em}
\end{figure*}

\begin{algorithm}[!t]
    \small
    \DontPrintSemicolon
\SetAlgoLined
\SetKwInOut{Input}{Input}
\SetKwInOut{Output}{Output}
\SetKwFunction{Tag}{Tag}
\SetKwFunction{Select}{SelectSkills}
\SetKwFunction{Gen}{GenerateSkill}
\SetKwFunction{Eval}{Evaluate}
\SetKwFunction{Fone}{TagF1}
\SetKwFunction{Top}{Top}
\SetKw{Return}{return}

\Input{
Initial skill library $\mathcal{L}_0$;
training task set $\mathcal{X}$;
number of evolution epochs $R$;
tag matching threshold $\delta$;
keep ratio $\rho$.
}
\Output{Evolved skill library $\mathcal{L}_R$.}

\BlankLine
Initialize $\mathcal{K}_s \leftarrow \Tag(s)$ for each $s\in\mathcal{L}_0$.\;

\For{$r \leftarrow 0$ \KwTo $R-1$}{
    $\mathcal{C}_r \leftarrow \emptyset$\;
    $\mathcal{F}_r \leftarrow \Eval(\mathcal{X}, \mathcal{L}_r)$
    \tcp*{failed pairs}

    \ForEach{$(x,\tau)\in\mathcal{F}_r$}{
        $\mathcal{K}^{\star} \leftarrow \Tag(x,\tau)$
        \tcp*{target tags}

        $S^{\star}\leftarrow
        \Select(\mathcal{L}_r,\mathcal{K}^{\star};\delta)$
        \tcp*{tag-aware source skills}

        $\mathcal{K}_{S^{\star}}\leftarrow
        \bigcup_{s\in S^{\star}}\mathcal{K}_s$\;

        $\mathcal{K}^{\star}_{\mathrm{un}}\leftarrow
        \mathcal{K}^{\star}\setminus\mathcal{K}_{S^{\star}}$
        \tcp*{uncovered tags}

        $c\leftarrow
        \Gen(x,\tau,S^{\star},\mathcal{K}^{\star}_{\mathrm{un}})$
        \tcp*{candidate skill}

        $\mathcal{K}_c\leftarrow\Tag(c)$\;

        $F_{\mathrm{tag}}(c)\leftarrow
        \Fone(\mathcal{K}_c,\mathcal{K}^{\star};\delta)$
        \tcp*{knowledge-tag coverage}

        $A(c)\leftarrow
        \sigma\!\left(
        \frac{1}{|c|}
        \left[
        \log \hat{p}(c\mid x)
        -
        \log \hat{p}(c\mid\emptyset)
        \right]\right)$
        \tcp*{task-only alignment}

        $\mathrm{Score}(c)\leftarrow
        \sqrt{F_{\mathrm{tag}}(c)A(c)}$\;

        $\mathcal{C}_r\leftarrow \mathcal{C}_r\cup\{c\}$\;
    }

    $\widetilde{\mathcal{C}}_r\leftarrow
    \Top_{\rho}(\mathcal{C}_r,\mathrm{Score})$
    \tcp*{retain top}

    $\mathcal{L}_{r+1}\leftarrow
    \mathcal{L}_r\cup\widetilde{\mathcal{C}}_r$\;

    Store $\mathcal{K}_c$ for each $c\in\widetilde{\mathcal{C}}_r$.\;
}

\Return{$\mathcal{L}_R$}\;
    \caption{\ourmethod}
    \label{alg:method_algorithm}
\end{algorithm}

In this section, we introduce \ourmethod, a skill evolution framework that improves skill libraries by increasing task-relevant knowledge coverage and selecting skills that are better aligned with target tasks.
At the evolution epoch $r$, we consider the following failed task–trajectory pairs:
\[
\mathcal{F}_r =
\{(x,\tau) \mid (x,\tau,q(x,\tau)) \in \mathcal{E}_r,\ q(x,\tau)=0\}.
\]
We focus on failed pairs because successful trajectories indicate that the current skill library already provides sufficient guidance for those tasks.
Therefore, to balance effectiveness and cost, \ourmethod evolves skills only from failures.

For each failed pair $(x,\tau)$, \ourmethod first infer task-relevant knowledge tags and retrieve skills with matching knowledge from the current library $\mathcal{L}_r$.
It then generates candidate skills conditioned on the selected skills, the failure trajectory, and the uncovered knowledge tags.
Finally, \ourmethod filters candidates using both knowledge-tag coverage and likelihood-based task alignment, and adds the retained skills to the library.
Figure~\ref{fig:method} and Algorithm~\ref{alg:method_algorithm} summarize the overall workflow.
All prompts are in Appendix~\ref{app:prompt}.

\subsection{Knowledge-Tag-Guided Source Skill Selection}

    We first retrieve source skills that provide knowledge relevant to the failed task.
    Specifically, each skill is associated with a set of \emph{knowledge tags}\footnote{We verify the reliability of LLM-generated knowledge tags in Appendix~\ref{app:tag_reliability}.}, where each tag is a short phrase describing the procedural knowledge contained in the skill, such as \texttt{video editing} or \texttt{PDF text extraction}.
    Before evolution, we generate a tag set $\mathcal{K}_{s_i}$ for each initial skill $s_i \in \mathcal{L}_0$.
    Whenever a new skill is added, its tag set is stored together for future epochs.
    
    Given a task-failure pair $(x,\tau)$, we use an LLM to infer the target knowledge tag set $\mathcal{K}^{\star}(x,\tau)$, which describes the knowledge required by the task.
    For a subset of source skills $S \subseteq \mathcal{L}_r$, we define the union of their knowledge tags as
    \[
        \mathcal{K}_S = \bigcup_{s \in S} \mathcal{K}_s .
    \]
    We select source skills by first maximizing the coverage of target tags and then, among subsets with the same coverage, minimizing irrelevant tags:
    \begin{equation}\label{equ:sample_optimize}
    \begin{gathered}
        \mathcal{S}_{\max}(x,\tau)
        =
        \argmax_{S \subseteq \mathcal{L}_r}
        \left|
            \mathcal{K}^{\star}(x,\tau) \cap \mathcal{K}_S
        \right|, \\
        S^{\star}(x,\tau)
        =
        \argmin_{S \in \mathcal{S}_{\max}(x,\tau)}
        \left|
            \mathcal{K}_S \setminus \mathcal{K}^{\star}(x,\tau)
        \right|.
    \end{gathered}
    \end{equation}
    This objective encourages the selected skills to transfer sufficient task-relevant knowledge while suppressing unrelated information.
    
    We further identify the target knowledge tags not covered by the selected skills using:
    \[
        \mathcal{K}^{\star}_{\mathrm{un}}(x,\tau;S^{\star})
        =
        \mathcal{K}^{\star}(x,\tau)
        \setminus
        \mathcal{K}_{S^{\star}} ,
    \]
    which serve as guidance for subsequent skill generation.
    Since LLM-generated tags may express the same concept with different surface forms, we embed all tags and treat two tags as equivalent when their cosine similarity is above $0.9$, following \citet{kim2025chopping}.
    Details of the source skill selection algorithm are provided in Appendix~\ref{app:sample_source_skill}.

\subsection{Knowledge-Tag-Guided Skill Evolution}

    After selecting source skills, \ourmethod uses an LLM to generate candidate skills.
    For each failed pair $(x,\tau) \in \mathcal{F}_r$, the LLM takes as input the task description $x$, the failure trajectory $\tau$, the selected source skills $S^{\star}(x,\tau)$, and the uncovered tags $\mathcal{K}^{\star}_{\mathrm{un}}(x,\tau;S^{\star})$.
    The LLM is prompted to transfer useful knowledge from the selected skills, cover missing knowledge when possible, and produce a reusable skill that can address the observed failure.
    
    To avoid overfitting to a single failed instance, the generated skill is required to generalize to a class of related tasks rather than only to describe the current trajectory and the task.
    All generated skills are collected into the candidate set $\mathcal{C}_r$.

\subsection{Skill Filtering by Knowledge Coverage and Task Alignment}

    Candidate skills generated by the previous step may still be incomplete, noisy, or weakly related to the target task.
    Therefore, \ourmethod filters $\mathcal{C}_r$ using two complementary criteria: knowledge-tag coverage and likelihood-based task alignment.
    
    For each candidate skill $c \in \mathcal{C}_r$, let $(x_c,\tau_c)$ denote the failed pair that produces it.
    We first generate the tag set $\mathcal{K}_c$ for $c$ and compare it with the target tag set $\mathcal{K}^{\star}(x_c,\tau_c)$.
    The precision and recall of knowledge coverage are defined as
    \[
    P_{\mathrm{tag}}(c) =
    \frac{
    \left|\mathcal{K}_c \cap \mathcal{K}^{\star}(x_c,\tau_c)\right|
    }{
    \left|\mathcal{K}_c\right|
    },
    \]
    \[
    R_{\mathrm{tag}}(c) =
    \frac{
    \left|\mathcal{K}_c \cap \mathcal{K}^{\star}(x_c,\tau_c)\right|
    }{
    \left|\mathcal{K}^{\star}(x_c,\tau_c)\right|
    }.
    \]
    We compute the knowledge coverage score based on the two values above as their F1 score:
    \[
    F_{\mathrm{tag}}(c)
    =
    \frac{
    2 P_{\mathrm{tag}}(c) R_{\mathrm{tag}}(c)
    }{
    P_{\mathrm{tag}}(c) + R_{\mathrm{tag}}(c)
    }.
    \]
    
    We then estimate whether the candidate skill is specifically aligned with the target task.
    Instead of relying on trajectory-centered filtering, which may favor skills that merely explain the observed execution, we measure task alignment by the conditional likelihood of the skill given the task description:
    \[
        A(c)=\sigma\left(
        \frac{1}{|c|}
        \left(
        \log{\hat{p}(c \mid x_c)}
        -
        \log{\hat{p}(c \mid \emptyset)}
        \right)
        \right),
    \]
    where $|c|$ is the token length of $c$ and $\sigma(\cdot)$ is the sigmoid function.
    The task-conditioned likelihood $\log{\hat{p}(c \mid x_c)}$ measures how well the skill is supported by the target task, calculated as the log probability of generating $c$ taken $x_c$ as input.
    Subtract the unconditional likelihood $\log{\hat{p}(c \mid \emptyset)}$ removes the model's prior preference for generally plausible but task-agnostic skills.
    Thus, $A(c)$ assigns higher scores to skills whose content is specifically grounded in the task requirements.

    Finally, we combine knowledge coverage and task alignment using their geometric mean:
    \[
        \mathrm{Score}(c)
        =
        \sqrt{
        F_{\mathrm{tag}}(c) A(c)
        }.
    \]
    This scoring function rewards candidates that perform well on both dimensions \cite{kubat-etal-1997-addressing}.
    We retain the top-ranked candidates as $\widetilde{\mathcal{C}}_r$ and update the skill library as
    \[
        \mathcal{L}_{r+1}
        =
        \mathcal{L}_r
        \cup
        \widetilde{\mathcal{C}}_r.
    \]

    \section{Experiment}
        \subsection{Setup}
    \paragraph{Models}
        To fully evaluate the generality and effectiveness of \ourmethod, we instantiate the agent with four representative LLM backbones: \textsc{DeepSeek-V4-Flash}~\cite{deepseekai2026deepseekv4}, \textsc{Claude-Haiku-4.5}~\cite{frontier-claude-haiku-4-5}, \textsc{GPT-5.4-nano}~\cite{frontier-gpt-5-4}, and \textsc{GPT-5.4}~\cite{frontier-gpt-5-4}.
        These backbones differ in model family, access setting, and scale, allowing us to test whether the evolved skill libraries transfer robustly across heterogeneous agents.

    \paragraph{Benchmarks and Metrics}
        We evaluate \ourmethod on three mainstream agent-skill benchmarks: SRA-Bench~\cite{su2026skillretrievalaugmentationagentic}, SkillsBench (S-Bench)~\cite{li2026skillsbenchbenchmarkingagentskills}, and SkillLearnBench (SL-Bench)~\cite{zhong2026skilllearnbenchbenchmarkingcontinuallearning}.
        Appendix~\ref{app:benchmark} provides detailed descriptions of these benchmarks.
        To match realistic skill-use scenarios, we do not provide the oracle skill for each test instance.
        Instead, the agent retrieves relevant skills and uses them to solve the task.
        We report the official metric: accuracy for SRA-Bench and pass rate for S-Bench and SL-Bench.

    \paragraph{Baselines}
        We compare \ourmethod with three current representative SOTA skill-evolution baselines: CoEvoSkills~\cite{zhang2026coevoskills}, Trace2Skill~\cite{ni2026trace2skilldistilltrajectorylocallessons}, and SkillX~\cite{wang2026skillx}.
        These methods cover major skill-evolution paradigms, including verification-driven refinement, trajectory-based distillation, and hierarchical skill-library construction.
        Their strong reported results and complementary designs make them suitable references for evaluating skill evolution.
        We introduce them in detail in Appendix~\ref{app:baseline}.

    \paragraph{Implementation Details}
        We run our method for $3$ evolution epochs and report the resulting library sizes in Appendix~\ref{app:evolve_scale}.
        For semantic matching between knowledge tags, we use all-MiniLM-L6-v2~\cite{reimers-2019-sentence-bert} with the threshold defined in \S\ref{sec:method}.
        The log-likelihood terms in the task-alignment score are estimated with \textsc{Qwen3.6-27B}~\cite{qwen3.6-27b}, where Appendix~\ref{app:score_model_sensitivity} analyzes the sensitivity to this score model.
        In each epoch, we generate one candidate skill for each failed task–trajectory pair with temperature $0.0$ and a maximum length of $16{,}394$ tokens, and retain the top $20\%$ candidates according to the final skill score\footnote{We analyze the effect of these parameters in Appendix~\ref{app:experiment_factor_affect}.}.
        Following \citet{zhao-etal-2024-expel}, we use a $3$-fold protocol with $3$ runs to avoid data leakage: skills are evolved on two folds and evaluated on the held-out fold.
        All evaluations use the official benchmark frameworks, and skills are retrieved with BM25 for each test instance.
        All analytical experiments of \ourmethod use \textsc{GPT-5.4-nano} with human-labeled initial skills to reduce inference costs while preserving reliable trends.

\subsection{Main Experiment}
    \begin{table*}[!t]
        \centering
        \small
        \setlength{\tabcolsep}{3pt}
\begin{tabular}{ll|ccc|ccc}
    \toprule
    \multicolumn{2}{c}{\textbf{Setting}} &   \multicolumn{3}{c}{\textbf{From-Scratch}} & \multicolumn{3}{c}{\textbf{Human-Labeled}} \\
    \cmidrule(lr){1-2} \cmidrule(lr){3-5} \cmidrule(lr){6-8} 
    \textbf{LLM} & \textbf{Baseline} 
    & \textbf{SRA-Bench}
    & \textbf{S-Bench}
    & \textbf{SL-Bench} 
    & \textbf{SRA-Bench}
    & \textbf{S-Bench}
    & \textbf{SL-Bench} \\
    \midrule

    \multirow{5}{*}{\textsc{DeepSeek-V4-Flash}}
    & Base          & $60.0$ & $13.1$ & $4.0$ & $68.9$ & $22.6$ & $21.0$ \\
    & CoEvoSkills   & $64.2$ & $23.8$ & $15.1$ & $72.6$ & $32.1$ & $30.0$\\
    & Trace2Skill   & $65.8$ & $21.4$ & $18.0$ & $73.7$ & $29.8$ & $33.0$\\
    & SkillX        & $65.3$ & $22.6$ & $16.0$ & $73.2$ & $31.0$ & $32.0$\\
    \cmidrule{2-8}
    & \ourmethod    & $\mathbf{67.8 \pm 0.7}$ & $\mathbf{26.2 \pm 1.7}$ & $\mathbf{22.0 \pm 1.6}$& $\mathbf{75.1 \pm 0.5}$ & $\mathbf{34.5 \pm 3.4}$ & $\mathbf{37.0 \pm 2.6}$  \\
    \midrule

     \multirow{5}{*}{\textsc{Claude-Haiku-4.5}}
     & Base          & $51.8$ & $19.0$ & $13.0$ & $60.2$ & $9.5$ & $20.0$\\
     & CoEvoSkills   & $57.9$ & $26.2$ & $21.0$ & $64.8$ & $28.6$ & $29.0$\\
     & Trace2Skill   & $59.3$ & $22.6$ & $24.1$ & $66.4$ & $23.8$ & $32.9$\\
     & SkillX        & $58.6$ & $25.0$ & $22.0$ & $65.7$ & $27.4$ & $31.0$\\
    \cmidrule{2-8}
     & \ourmethod    & $\mathbf{62.4 \pm 0.5}$ & $\mathbf{28.6 \pm 2.9}$ & $\mathbf{27.0 \pm 2.8}$ & $\mathbf{68.1 \pm 0.6}$ & $\mathbf{31.0 \pm 1.7}$ & $\mathbf{35.0 \pm 2.3}$\\
    \midrule

      \multirow{5}{*}{\textsc{GPT-5.4-nano}}
     & Base          & $43.6$ & $3.6$ & $17.0$ & $47.3$ & $9.5$ & $30.0$ \\
     & CoEvoSkills   & $50.3$ & $16.7$ & $24.0$  & $53.6$ & $25.0$ & $36.0$\\
     & Trace2Skill   & $52.1$ & $14.3$ & $26.1$ & $55.9$ & $22.6$ & $39.0$\\
     & SkillX        & $51.5$ & $15.5$ & $25.0$ & $54.9$ & $23.8$ & $38.1$\\
    \cmidrule{2-8}
     & \ourmethod    & $\mathbf{55.5 \pm 0.7}$ & $\mathbf{19.0 \pm 3.4}$ & $\mathbf{28.0 \pm 2.7}$ & $\mathbf{58.0 \pm 0.5}$ & $\mathbf{26.2 \pm 1.7}$ & $\mathbf{41.0 \pm 2.0}$ \\
    \midrule

      \multirow{5}{*}{\textsc{GPT-5.4}}
     & Base          & $64.1$ & $39.3$ & $42.0$  & $70.7$ & $45.2$ & $48.0$\\
     & CoEvoSkills   & $68.6$ & $46.4$ & $50.0$ & $74.9$ & $52.4$ & $56.0$ \\
     & Trace2Skill   & $69.8$ & $42.9$ & $52.0$  & $75.7$ & $50.0$ & $58.0$\\
     & SkillX        & $69.2$ & $45.2$ & $51.0$ & $75.3$ & $51.2$ & $57.0$\\
    \cmidrule{2-8}
     & \ourmethod    & $\mathbf{72.2 \pm 0.3}$ & $\mathbf{48.8 \pm 1.7}$ & $\mathbf{54.0 \pm 3.1}$  & $\mathbf{77.3 \pm 0.7}$ & $\mathbf{54.8 \pm 3.4}$ & $\mathbf{59.0 \pm 2.2}$ \\

    \bottomrule
\end{tabular}
        \caption{
            Main results across skill-evolution methods.
            \texttt{From-Scratch} denotes evolution without human-labeled initial skills, while \texttt{Human-Labeled} denotes evolution from the original human-labeled skill library.
            \texttt{Base} denotes direct reasoning in \texttt{From-Scratch} and retrieval from the original skill library in \texttt{Human-Labeled}.
            Except \texttt{Base}, all results are averaged on $3$-fold runs.
            The best result of each setting is marked in \textbf{bold}.
        }
        \vspace{-1em}
        \label{tab:main_experiment}
    \end{table*}

    Table~\ref{tab:main_experiment} shows that \ourmethod achieves the best result across all benchmarks, backbone models, and initialization settings.
    Compared with \texttt{Base}, \ourmethod improves average performance by $11.9$ points, corresponding to a relative gain of $\mathbf{34.7\%}$.
    It also consistently outperforms existing agent skill evolution baselines, exceeding the strongest prior method by $3.3$ points on average and establishing a new SOTA for skill evolution.
    Case studies of skills generated by \ourmethod are provided in Appendix~\ref{app:case_study}.
    We can also see that:

    \paragraph{Benchmark.}
        \ourmethod brings stable gains on all three experimental benchmarks.
        On SRA-Bench, where \texttt{Base} is already relatively strong, \ourmethod still yields a $15.9\%$ relative improvement, indicating that the evolved skills further improve skill utilization.
        On S-Bench, it achieves the largest relative gain of $134.8\%$, showing strong generalization and effectiveness to diverse real-world tasks.
        On SL-Bench, it obtains the largest absolute gain of $13.5$ points, demonstrating its effectiveness in skill evolution.

    \paragraph{Model.}
        \ourmethod improves all four backbone models, suggesting that its benefits are largely model-agnostic.
        The gains are especially notable for smaller models, including \textsc{DeepSeek-V4-Flash}, \textsc{Claude-Haiku-4.5}, and \textsc{GPT-5.4-nano}, where evolved skills provide valuable procedural guidance.
        For the stronger \textsc{GPT-5.4}, the relative improvement is smaller because of its higher \texttt{Base} performance, but \ourmethod still remains the best-performing method on this model, showing its effectiveness.

    \paragraph{Baseline.}
        The clear improvement over \texttt{Base} confirms the necessity of skill evolution.
        Compared with CoEvoSkills, Trace2Skill, and SkillX, \ourmethod improves average performance by $3.7$, $3.3$, and $3.4$ points, respectively.
        While prior methods exhibit benchmark-dependent strengths, \ourmethod achieves the best results in every setting and a new SOTA result, indicating more generally effective skill evolution.

\subsection{Ablation Study}
    \begin{table}[!t]
        \centering
        \small
        \begin{tabular}{l|ccc}
    \toprule
    \textbf{Method} & \textbf{SRA-Bench} & \textbf{S-Bench} & \textbf{SL-Bench} \\
    \midrule
    \ourmethod & $58.0$ & $25.8$ & $41.4$ \\
    - Sampling & $53.6$ & $18.2$ & $36.5$ \\
    - Generation & $54.1$ & $16.7$ & $35.4$ \\
    - Filter & $55.0$ & $20.3$ & $38.2$ \\
    \bottomrule
\end{tabular}

        \caption{
            Component ablation of \ourmethod.
            \textit{- Sampling} removes knowledge-tag-guided source-skill selection.
            \textit{- Generation} directly concatenates selected source skills without synthesizing a new candidate skill.
            \textit{- Filter} keeps all generated candidate skills.
        }
        \vspace{-1.5em}
        \label{tab:ablation_study}
    \end{table}

    We conduct ablation studies to isolate the contribution of each component in \ourmethod.
    As shown in Table~\ref{tab:ablation_study}, removing any component adversely affects performance.
    \textit{(i)} Removing knowledge-tag-guided sampling causes a $5.6$-point average drop, showing that selecting source skills based on task-required knowledge is important for effective generation.
    \textit{(ii)} Removing skill generation leads to the largest drop of $6.3$ points, indicating that directly reusing selected source skills is insufficient and that the agent must adapt and synthesize transferable knowledge according to the failed task–trajectory pair.
    \textit{(iii)} Removing filtering also reduces performance, confirming that post-generation control is needed to suppress noisy or weakly aligned skills before they enter the library.

\subsection{Evolution Efficiency across Baselines}
    \begin{figure}[!t]
        \centering
        \small
        \makebox[\linewidth][c]{%
\resizebox{\linewidth}{!}{%
\begin{tikzpicture}[
    font=\scriptsize,
    x=1cm,
    y=1cm,
    bar/.style={draw=none},
]

\path[use as bounding box] (-0.08,-0.62) rectangle (5.66,2.93);

\def\tokscale{0.00036}
\def\timescale{0.00098}
\def\bw{0.078}

\def\costbar#1#2#3#4#5#6{%
    \fill[bar, fill=#5]
        ({#1+#3-\bw},0) rectangle ({#1+#3+\bw},{#2*#4});
    \node[
        font=\tiny,
        text=#5,
        rotate=90,
        anchor=west,
        inner sep=0.15pt
    ] at ({#1+#3},{#2*#4+0.03}) {\scalebox{0.76}{#6}};
}

\begin{scope}[shift={(0,0)}]

\node[font=\scriptsize\bfseries] at (1.35,2.40) {Token Cost (K)};

\costbar{0.45}{348}{-0.27}{\tokscale}{softblue}{348}
\costbar{1.35}{4395}{-0.27}{\tokscale}{softblue}{4,395}
\costbar{2.25}{2981}{-0.27}{\tokscale}{softblue}{2,981}

\costbar{0.45}{399}{-0.09}{\tokscale}{softgreen}{399}
\costbar{1.35}{5038}{-0.09}{\tokscale}{softgreen}{5,038}
\costbar{2.25}{3417}{-0.09}{\tokscale}{softgreen}{3,417}

\costbar{0.45}{255}{0.09}{\tokscale}{softgray}{255}
\costbar{1.35}{3216}{0.09}{\tokscale}{softgray}{3,216}
\costbar{2.25}{2181}{0.09}{\tokscale}{softgray}{2,181}

\costbar{0.45}{223}{0.27}{\tokscale}{softred}{223}
\costbar{1.35}{2815}{0.27}{\tokscale}{softred}{2,815}
\costbar{2.25}{1909}{0.27}{\tokscale}{softred}{1,909}

\draw[gray!45] (-0.05,0) -- (2.75,0);
\node at (0.45,-0.18) {SRA};
\node at (1.35,-0.18) {S};
\node at (2.25,-0.18) {SL};

\end{scope}

\begin{scope}[shift={(2.86,0)}]

\node[font=\scriptsize\bfseries] at (1.35,2.40) {Time Cost (s)};

\costbar{0.45}{101.4}{-0.27}{\timescale}{softblue}{101.4}
\costbar{1.35}{1280.3}{-0.27}{\timescale}{softblue}{1280.3}
\costbar{2.25}{868.4}{-0.27}{\timescale}{softblue}{868.4}

\costbar{0.45}{146.0}{-0.09}{\timescale}{softgreen}{146.0}
\costbar{1.35}{1843.6}{-0.09}{\timescale}{softgreen}{1843.6}
\costbar{2.25}{1250.4}{-0.09}{\timescale}{softgreen}{1250.4}

\costbar{0.45}{93.3}{0.09}{\timescale}{softgray}{93.3}
\costbar{1.35}{1176.9}{0.09}{\timescale}{softgray}{1176.9}
\costbar{2.25}{798.1}{0.09}{\timescale}{softgray}{798.1}

\costbar{0.45}{81.6}{0.27}{\timescale}{softred}{81.6}
\costbar{1.35}{1030.1}{0.27}{\timescale}{softred}{1030.1}
\costbar{2.25}{698.7}{0.27}{\timescale}{softred}{698.7}

\draw[gray!45] (-0.05,0) -- (2.75,0);
\node at (0.45,-0.18) {SRA};
\node at (1.35,-0.18) {S};
\node at (2.25,-0.18) {SL};

\end{scope}

\begin{scope}[shift={(0.1,-0.50)}]

\fill[softblue]  (0,0) rectangle (0.12,0.07);
\node[anchor=west, font=\tiny] at (0.16,0.035) {CoEvoSkills};

\fill[softgreen] (1.43,0) rectangle (1.55,0.07);
\node[anchor=west, font=\tiny] at (1.59,0.035) {Trace2Skill};

\fill[softgray]  (2.92,0) rectangle (3.04,0.07);
\node[anchor=west, font=\tiny] at (3.08,0.035) {SkillX};

\fill[softred]   (3.86,0) rectangle (3.98,0.07);
\node[anchor=west, font=\tiny] at (4.02,0.035) {\ourmethod};

\end{scope}

\end{tikzpicture}%
}%
}
        \vspace{-2em}
        \caption{
            Token and wall-clock costs of skill evolution on each benchmark across various methods.
            The lowest cost under each setting is marked in \textbf{bold}.
        }
        \label{fig:token_cost_cross_baseline}
    \end{figure}

    Figure~\ref{fig:token_cost_cross_baseline} compares the evolution cost of \ourmethod with existing skill-evolution baselines.
    \ourmethod achieves the lowest token and wall-clock cost on all three benchmarks while also producing the strongest downstream performance.
    Thus, its gains do not come from a larger evolution budget.
    Instead, \ourmethod offers a better cost–performance trade-off.

    Besides, we can make three observations.
    \textit{(i)} \ourmethod is consistently more efficient than prior baselines.
    Compared with the average cost of existing skill-evolution methods, it reduces token usage by about $33.2\%$ and time cost by about $28.1\%$.
    Even compared with the cheapest baseline, SkillX, \ourmethod still saves about $12.5\%$ tokens and time on average.
    \textit{(ii)} These savings become more important as benchmark scale increases.
    For example, compared with Trace2Skill, \ourmethod saves $2{,}223$K tokens and $813.5$s on S-Bench, and $1{,}508$K tokens and $551.7$s on SL-Bench.
    This suggests that targeted skill evolution scales better than broader trajectory distillation or library-construction pipelines.
    \textit{(iii)} Higher evolution costs do not necessarily produce better skills.
    Trace2Skill consumes the most tokens and time, but it does not achieve the best downstream performance.
    These results indicate that effective evolution should improve the utility of each generated skill rather than merely generating or refining more content.
    By selecting source skills with task-relevant knowledge tags and filtering candidates through coverage and alignment scores, \ourmethod avoids many redundant or weakly useful evolution steps, leading to both stronger performance and lower costs.

\subsection{Performance Across Knowledge Coverage and Task Alignment}
    \begin{figure}[!t]
        \centering
        \begin{tikzpicture}

\begin{axis}[
    width=\linewidth,
    height=0.618\linewidth,
    xlabel={Knowledge-tag coverage score $F_{\mathrm{tag}}$},
    ylabel={Task-alignment score $A$},
    xmin=0.42, xmax=0.61,
    ymin=0.62, ymax=0.84,
    xtick={0.44,0.48,0.52,0.56,0.60},
    ytick={0.65,0.70,0.75,0.80},
    tick label style={font=\scriptsize},
    label style={font=\scriptsize},
    grid=both,
    grid style={line width=.1pt, draw=gray!18},
    major grid style={line width=.2pt, draw=gray!30},
    axis line style={black!70},
    tick style={black!70},
    clip=false,
]

\draw[->, line width=0.7pt, draw=gray!55]
    (axis cs:0.445,0.655) -- (axis cs:0.570,0.802);
\node[font=\tiny, text=gray!65, anchor=west]
    at (axis cs:0.505,0.805) {higher scores};

\addplot[
    only marks,
    mark=*,
    mark size=3pt,
    draw=softgray,
    fill=softgray!55
] coordinates {(0.438,0.645)};
\node[font=\tiny, anchor=north east, xshift=-1pt, yshift=-1pt]
    at (axis cs:0.438,0.645) {34.3};

\addplot[
    only marks,
    mark=square*,
    mark size=3pt,
    draw=softblue,
    fill=softblue!65
] coordinates {(0.482,0.723)};
\node[font=\tiny, anchor=east, xshift=-3pt, yshift=3pt]
    at (axis cs:0.482,0.723) {42.5};

\addplot[
    only marks,
    mark=triangle*,
    mark size=3pt,
    draw=softgreen,
    fill=softgreen!65
] coordinates {(0.490,0.738)};
\node[font=\tiny, anchor=west, xshift=4pt, yshift=3pt]
    at (axis cs:0.490,0.738) {42.9};

\addplot[
    only marks,
    mark=diamond*,
    mark size=3pt,
    draw=black!65,
    fill=black!40
] coordinates {(0.525,0.705)};
\node[font=\tiny, anchor=north west, xshift=3pt, yshift=-2pt]
    at (axis cs:0.525,0.705) {42.8};

\addplot[
    only marks,
    mark=*,
    mark size=3pt,
    draw=softred!90!black,
    fill=softred!75
] coordinates {(0.577,0.808)};
\node[
    font=\tiny\bfseries,
    anchor=west,
    xshift=4pt,
    text=softred!80!black
] at (axis cs:0.577,0.808) {46.2};

\end{axis}

\node[font=\tiny, anchor=north] at (current bounding box.south) {%
    \tikz{\draw[softgray, fill=softgray!55] (0,0) circle (1.5pt);}~Base
    \quad
    \tikz{\draw[softblue, fill=softblue!65] (-1.5pt,-1.5pt) rectangle (1.5pt,1.5pt);}~CoEvoSkills
    \quad
    \tikz{\draw[softgreen, fill=softgreen!65] (0,1.8pt) -- (-1.8pt,-1.4pt) -- (1.8pt,-1.4pt) -- cycle;}~Trace2Skill
    \quad
    \tikz{\draw[black!65, fill=black!40] (0,2pt) -- (2pt,0) -- (0,-2pt) -- (-2pt,0) -- cycle;}~SkillX
    \quad
    \tikz{\draw[softred!90!black, fill=softred!75] (0,0) circle (1.5pt);}~\ourmethod
};

\end{tikzpicture}
        \vspace{-0.6em}
        \caption{
            Joint relation between knowledge-tag coverage, task alignment, and downstream performance averaged across benchmarks.
            The $x$- and $y$-axes show the averaged scores across SRA-Bench, S-Bench, and SL-Bench, while the number beside each point denotes the average performance.
        }
        \label{fig:coverage_alignment_performance}
    \end{figure}

    \begin{table*}[!t]
        \centering
        \small
        \begin{tabular}{l|cccc|cccc}
    \toprule
    \multirow{2}{*}{\textbf{Method}} 
    & \multicolumn{4}{c|}{\textbf{Source-Skill Sampling}} 
    & \multicolumn{4}{c}{\textbf{Candidate-Skill Filtering}} \\
    & \textbf{SRA} & \textbf{S} & \textbf{SL} & \textbf{Avg.}
    & \textbf{SRA} & \textbf{S} & \textbf{SL} & \textbf{Avg.} \\
    \midrule
    BM25 
    & $54.9$ & $20.7$ & $37.9$ & $37.8$
    & $55.6$ & $21.6$ & $39.1$ & $38.8$ \\
    Embedding 
    & $55.6$ & $21.8$ & $38.8$ & $38.7$
    & $56.4$ & $23.1$ & $39.9$ & $39.8$ \\
    LLM-Based 
    & $56.7$ & $23.5$ & $40.2$ & $40.1$
    & $57.0$ & $24.3$ & $40.5$ & $40.6$ \\
    \ourmethod 
    & $\mathbf{58.0}$ & $\mathbf{25.8}$ & $\mathbf{41.4}$ & $\mathbf{41.7}$
    & $\mathbf{58.0}$ & $\mathbf{25.8}$ & $\mathbf{41.4}$ & $\mathbf{41.7}$ \\
    \bottomrule
\end{tabular}
        \caption{
            Performance of \ourmethod under different source-skill sampling and candidate-skill filtering strategies.
            We replace the sampling and filtering methods of \ourmethod.
            SRA, S, and SL denote SRA-Bench, S-Bench, and SL-Bench, respectively.
            The best result within each strategy group is marked in \textbf{bold}.
        }
        \vspace{-1em}
        \label{tab:sample_filter_with_different_method}
    \end{table*}

    In this paper, we further examine whether knowledge coverage and task alignment are associated with downstream skill quality, and whether \ourmethod improves both dimensions simultaneously.
    Figure~\ref{fig:coverage_alignment_performance} plots each method by its averaged knowledge-tag coverage score $F_{\mathrm{tag}}$, task-alignment score $A$, and downstream performance across benchmarks, settings, and backbone models.

    The results show a clear positive trend: methods with higher $F_{\mathrm{tag}}$ and $A$ generally achieve better task-solving performance.
    \texttt{Base} appears in the lower-left region and obtains the weakest result, while skill-evolution baselines move toward higher coverage and alignment, improving downstream performance.
    \ourmethod lies in the upper-right region, achieving the best averaged $F_{\mathrm{tag}}$, the best averaged $A$, and the highest averaged performance.
    Compared with the strongest baseline, it improves task-relevant knowledge coverage by $9.9\%$ and task-alignment likelihood by $8.9\%$.
    This supports our motivation, where effective skill evolution should not only retrieve or synthesize broadly relevant skills but also ensure that evolved skills cover the knowledge required by the task and remain aligned with the task objective.

    The baseline comparison further shows that knowledge coverage and task alignment are complementary.
    SkillX achieves relatively strong knowledge-tag coverage among baselines, whereas Trace2Skill obtains stronger task alignment, yet their downstream performance is similar.
    This suggests that improving only one dimension can create a bottleneck.
    By optimizing both dimensions jointly, \ourmethod produces more balanced and task-useful skills, which explains its consistent advantage over prior skill-evolution methods.

\subsection{Effect of Sampling and Filtering Method}
    We further analyze two key decisions in skill evolution: selecting source skills before generation and filtering generated candidates afterward.
    These two stages affect skill quality from different directions: source-skill sampling determines what prior knowledge is transferred into generation, while candidate-skill filtering controls which generated skills are finally added to the library.
    The corresponding experimental results are shown in Table~\ref{tab:sample_filter_with_different_method}.

    For source-skill sampling, the average score increases from $37.8$ with BM25 to $38.7$ with Embedding, $40.1$ with LLM-Based selection, and $41.7$ with \ourmethod.
    This trend shows that the quality of retrieved source skills directly affects the generated skills.
    Lexical matching and generic semantic similarity can miss fine-grained procedural knowledge, whereas knowledge-tag-guided sampling better identifies the missing knowledge required by failed tasks.
    By maximizing target-tag coverage while reducing irrelevant tags, \ourmethod provides less noisy evidence for the skill evolution.

    For candidate-skill filtering, the average score improves from $38.8$ to $41.7$.
    This confirms that post-generation filtering is also important for building a useful skill library.
    However, candidate filtering is not simply a relevance-ranking problem.
    By jointly considering knowledge coverage and likelihood-based task alignment, \ourmethod removes superficially related but weakly useful candidates and retains skills that better support downstream task solving.
    This complements source skill sampling: even when relevant source skills are selected, generated candidates may still overfit the trajectory, omit key knowledge, or include task-agnostic procedures.
    The gains are especially clear for both sampling and filtering, suggesting that precise knowledge transfer and alignment-aware filtering are most valuable when tasks involve diverse execution constraints.

    \section{Conclusion}
        In this paper, we propose \ourmethod, a knowledge-aware and task-aligned framework for agent skill evolution. 
        \ourmethod uses knowledge tags to identify task-required procedural knowledge, retrieve complementary source skills, and generate candidate skills for failed tasks. 
        It further filters evolved skills by combining knowledge-tag coverage with likelihood-based task alignment, retaining skills that are both informative and task-relevant while reducing noisy or incomplete transfer. 
        Experiments on $3$ benchmarks with $4$ LLM backbones show that \ourmethod consistently improves agent performance, achieving an average gain of $11.9$ points over \texttt{Base} and outperforming existing skill-evolution baselines. 
        Ablation studies and diagnostic analyzes confirm the importance of knowledge-guided sampling, adaptive skill generation, and alignment-aware filtering in building higher-quality skill libraries. 
        Overall, these results show that effective skill evolution should explicitly model both the knowledge required and the alignment to the target task.

    \clearpage
    \section*{Limitations}
        While \ourmethod improves skill evolution through knowledge-tag-guided sampling and alignment filtering, our study has several limitations. 
        First, our experiments are conducted on three skill-oriented benchmarks and four LLM backbones. 
        These settings are diverse, but they may not cover all real-world agent scenarios. 
        Besides, our method benefits from informative failure trajectories and reliable task feedback, and noisy traces may lead to less useful evolved skills. 
        Future work can extend \ourmethod to broader environments and explore stronger tag normalization, verification, and alignment scoring.
    \section*{Ethics Statement}
        This work aims to improve reusable skill evolution for benign LLM-agent tasks. 
        It does not involve private user data or new human-subject annotation.
        The usage of benchmarks and LLMs follows their licenses.
        We employ the LLM tool to polish the paper writing.

    \bibliography{custom}

    \clearpage
    \appendix
    \section{Prompt}\label{app:prompt}
    \begin{figure*}[!t]
        \centering
        \small
        \input{tab/prompt/tag_with_skill}
        \caption{
            Prompt for generating knowledge tags from a given skill.
        }
        \label{fig:prompt_tag_with_skill}
    \end{figure*}
    \begin{figure*}[!t]
        \centering
        \small
        \input{tab/prompt/tag_with_task_trace}
        \caption{
            Prompt for inferring target knowledge tags from a task-trajectory pair.
        }
        \label{fig:prompt_tag_with_task_trace}
    \end{figure*}
    \begin{figure*}[!t]
        \centering
        \small
        \input{tab/prompt/generate_skill}
        \caption{
            Prompt for synthesizing a candidate skill during skill evolution.
        }
        \label{fig:prompt_generate_skill}
    \end{figure*}

    Figure~\ref{fig:prompt_tag_with_skill}, Figure~\ref{fig:prompt_tag_with_task_trace}, and Figure~\ref{fig:prompt_generate_skill} show the prompts used in \ourmethod for skill-level tag generation, task-trajectory-level target tag generation, and candidate-skill synthesis, respectively.

\section{Algorithm of Sampling Source Skill}\label{app:sample_source_skill}
    \begin{table}[!t]
        \centering
        \small
        \begin{tabular}{l|ccc}
    \toprule
    \textbf{Method} & \textbf{SRA-Bench} & \textbf{S-Bench} & \textbf{SL-Bench} \\
    \midrule
    Greedy & $58.0$ & $25.8$ & $41.4$ \\
    Primal-Dual & $57.3$ & $25.2$ & $40.6$ \\
    LP Relaxation & $58.2$ & $25.7$ & $41.8$ \\
    \bottomrule
\end{tabular}
        \caption{
            Performance of \ourmethod with different source-skill sampling algorithms.
        }
        \label{tab:performance_with_different_sample_algorithm}
    \end{table}

    The source-skill selection objective in Equation~\ref{equ:sample_optimize} can be regarded as a variant of the minimum set cover problem, where the goal is to cover as many target knowledge tags as possible while introducing as few irrelevant tags as possible.
    Since the current skill library may not contain skills covering every target tag, all algorithms first operate on the coverable subset of $\mathcal{K}^{\star}$ and keep the remaining uncovered tags for subsequent candidate-skill generation.
    Table~\ref{tab:performance_with_different_sample_algorithm} compares three sampling algorithms.
    Because their performance differences are small, \ourmethod adopts greedy search by default for lower computational cost.

    \paragraph{Greedy Search.}
        Greedy search builds the source-skill subset iteratively.
        Starting from an empty subset, it repeatedly selects the skill that provides the largest marginal coverage over the currently uncovered target tags.
        In our setting, the benefit is the number of newly covered tags in $\mathcal{K}^{\star}$, and the cost is the number of newly introduced tags outside $\mathcal{K}^{\star}$.
        Thus, the algorithm prefers skills that cover more missing task-relevant knowledge while adding less unrelated knowledge.
        The selection stops when all coverable target tags are covered or no remaining skill can increase coverage.
        We then remove redundant skills whose deletion does not reduce the covered target-tag set.

    \paragraph{Primal-Dual Approximation.}
        We formulate each target knowledge tag as an element to be covered and each source skill as a candidate set with a cost.
        The cost of a skill is determined by the irrelevant tags it introduces, i.e., tags in $\mathcal{K}_{s_i} \setminus \mathcal{K}^{\star}$.
        The algorithm maintains dual variables for uncovered target tags and increases them until at least one skill becomes cost-effective.
        That skill is selected, and the newly covered tags are removed from the uncovered set.
        This process continues until all coverable target tags are covered.
        By incorporating irrelevant-tag cost, the primal-dual algorithm follows the same principle as \ourmethod: improving knowledge coverage while reducing noisy knowledge passed to skill generation.

    \paragraph{LP Relaxation with Rounding.}
        LP relaxation casts source-skill sampling as a relaxed integer optimization problem.
        Each skill has a selection variable, and each irrelevant tag has an auxiliary variable indicating whether it is introduced by the selected skills.
        The coverage constraints require every coverable target tag in $\mathcal{K}^{\star}$ to be covered by at least one selected skill, while the objective minimizes the number of introduced irrelevant tags.
        Instead of solving the binary program directly, we relax the selection variables to continuous values in $[0,1]$.
        The final source-skill subset is obtained by rounding high-valued skills and, when necessary, applying a greedy repair step to satisfy any remaining coverage constraints.

\section{Experimental Benchmark}\label{app:benchmark}
    \paragraph{SRA-Bench.}
        SRA-Bench evaluates skill retrieval augmentation for LLM-based agents.
        Rather than assuming that the oracle skill is directly provided, it studies a realistic setting where an agent must retrieve relevant skills from a large external corpus and use them for task solving.
        The benchmark decomposes evaluation into skill retrieval, skill incorporation, and end-task execution, allowing fine-grained analysis of agent failures.
        In our experiments, SRA-Bench is used to evaluate whether an evolved skill library helps the agent retrieve and use task-relevant skills for downstream execution.

    \paragraph{SkillsBench.}
        SkillsBench, abbreviated as S-Bench in our experiments, measures how effectively LLM agents use external skills across diverse real-world tasks.
        Each task is paired with curated skills and deterministic verifiers, making it possible to compare agents under different skill-use settings, such as no skills, curated skills, and self-generated skills.
        Therefore, S-Bench is suitable for evaluating both the utility of high-quality expert skills and the practical limitations of automatically generated skills.
        We use S-Bench to test whether retrieved skills from the current library can improve task completion when the target skill is not given in advance.

    \paragraph{SkillLearnBench.}
        SkillLearnBench, abbreviated as SL-Bench, focuses on continual skill learning from agent experience.
        It evaluates skill-dependent real-world tasks and considers not only final task success but also the quality of learned skills and execution trajectories.
        Unlike benchmarks that only test the use of fixed skills, SL-Bench emphasizes whether an agent can acquire, refine, and reuse skills across tasks.
        We use SL-Bench to examine whether the evolved skill library can provide reusable procedural guidance for future task solving.

\section{Experimental Baseline}\label{app:baseline}
    \paragraph{CoEvoSkills.}
        CoEvoSkills is a verification-driven skill evolution framework based on a generate-verify-refine loop~\cite{zhang2026coevoskills}.
        It couples a skill generator with a surrogate verifier, which provides actionable feedback without requiring access to ground-truth test cases.
        This iterative design enables agents to improve complex skill packages beyond one-shot generation.
        We include CoEvoSkills as a representative baseline for verification-guided skill refinement.

    \paragraph{Trace2Skill.}
        Trace2Skill distills transferable skills from execution trajectories~\cite{ni2026trace2skilldistilltrajectorylocallessons}.
        Instead of relying on a single trajectory, it analyzes diverse trajectories with multiple sub-agents, extracts trajectory-local lessons, and consolidates them into a unified skill directory through inductive reasoning.
        It supports both creating new skills from scratch and refining existing human-written skills.
        We use Trace2Skill as a representative trajectory-driven skill distillation baseline.

    \paragraph{SkillX.}
        SkillX automatically constructs reusable skill knowledge bases from agent experience~\cite{wang2026skillx}.
        It organizes experience into a multi-level hierarchy of planning skills, functional skills, and atomic skills, and further improves the library through iterative refinement and exploratory expansion.
        The resulting skill base is designed to be plug-and-play across agents and environments.
        We compare with SkillX as a representative structured skill-library construction baseline.

\section{Evolved Scale of \ourmethod}\label{app:evolve_scale}
    \begin{table}[!t]
        \centering
        \small
        \resizebox{\linewidth}{!}{
\begin{tabular}{l|ccc}
    \toprule
    \textbf{Model} & \textbf{SRA-Bench} & \textbf{S-Bench} & \textbf{SL-Bench} \\
    \midrule
    \textsc{DeepSeek-V4-Flash} & $1{,}140$ & $31$ & $28$ \\
    \textsc{Claude-Haiku-4.5} & $984$ & $27$ & $25$ \\
    \textsc{GPT-5.4-nano} & $1{,}036$ & $29$ & $26$ \\
    \textsc{GPT-5.4} & $829$ & $23$ & $21$ \\
    \bottomrule
\end{tabular}
}
        \caption{
            Number of skills generated by \ourmethod under each setting.
        }
        \label{tab:evolved_scale}
    \end{table}

    Table~\ref{tab:evolved_scale} reports the scale of the evolved skill libraries produced by \ourmethod under different models and benchmarks.

\section{Reliability of Knowledge Tags Generated by LLMs}\label{app:tag_reliability}
    Since \ourmethod uses LLM-generated knowledge tags for both source-skill selection and candidate-skill filtering, we examine whether these tags reliably capture reusable procedural knowledge.
    We conduct a human-audited study over two tag sources: tags generated from existing skill descriptions and target tags inferred from failed task-trajectory pairs.
    For each benchmark, we sample $60$ skill descriptions and $60$ failed task-trajectory pairs, resulting in $360$ evaluated items in total.
    Two annotators assign capability-level reference tags to each item, and disagreements are resolved through discussion.

    We evaluate tag quality with exact-match F1, semantic precision, semantic recall, semantic F1, and self-consistency.
    Exact-match F1 requires normalized tag strings to be identical.
    Semantic matching instead treats two tags as equivalent when their embedding cosine similarity exceeds the same threshold $\delta=0.9$ used in \ourmethod.
    Self-consistency is computed as the average pairwise semantic F1 across three independent tag-generation runs.

    \begin{figure}[!t]
        \centering
        \resizebox{\linewidth}{!}{%
\begin{tikzpicture}

\definecolor{tagblue}{RGB}{52,112,181}
\definecolor{tagnavy}{RGB}{25,54,97}

\newcommand{\relcell}[5]{
    \node[
        minimum width=1.35cm,
        minimum height=0.54cm,
        rounded corners=2pt,
        fill=tagblue!#4,
        draw=white,
        line width=0.8pt,
        font=\scriptsize\bfseries,
        text=#5
    ] at (#1,#2) {#3};
}

\node[font=\scriptsize\bfseries, text=tagnavy] at (-1.15,0.75) {Tag Source};
\node[font=\scriptsize\bfseries, text=tagnavy] at (0.25,0.75) {Exact F1};
\node[font=\scriptsize\bfseries, text=tagnavy] at (1.65,0.75) {Sem. P};
\node[font=\scriptsize\bfseries, text=tagnavy] at (3.05,0.75) {Sem. R};
\node[font=\scriptsize\bfseries, text=tagnavy] at (4.45,0.75) {Sem. F1};
\node[font=\scriptsize\bfseries, text=tagnavy] at (5.85,0.75) {Self-F1};

\node[anchor=east, font=\scriptsize, text=black!80] at (-0.55,0.05) {Skill};
\node[anchor=east, font=\scriptsize, text=black!80] at (-0.55,-0.65) {Task--trace};
\node[anchor=east, font=\scriptsize\bfseries, text=tagnavy] at (-0.55,-1.35) {Overall};

\relcell{0.25}{0.05}{$0.74$}{46}{black}
\relcell{1.65}{0.05}{$0.86$}{72}{white}
\relcell{3.05}{0.05}{$0.80$}{58}{white}
\relcell{4.45}{0.05}{$0.83$}{66}{white}
\relcell{5.85}{0.05}{$0.88$}{78}{white}

\relcell{0.25}{-0.65}{$0.70$}{36}{black}
\relcell{1.65}{-0.65}{$0.82$}{64}{white}
\relcell{3.05}{-0.65}{$0.77$}{52}{white}
\relcell{4.45}{-0.65}{$0.79$}{56}{white}
\relcell{5.85}{-0.65}{$0.84$}{68}{white}

\relcell{0.25}{-1.35}{$0.72$}{41}{black}
\relcell{1.65}{-1.35}{$0.84$}{68}{white}
\relcell{3.05}{-1.35}{$0.79$}{56}{white}
\relcell{4.45}{-1.35}{$\mathbf{0.81}$}{62}{white}
\relcell{5.85}{-1.35}{$\mathbf{0.86}$}{72}{white}

\draw[
    tagnavy,
    line width=0.8pt,
    rounded corners=3pt
] (3.75,-1.66) rectangle (6.55,-1.04);

\node[anchor=west, font=\tiny, text=black!65] at (-0.35,-2.23) {Lower};
\foreach \i/\shade in {0/30,1/42,2/54,3/66,4/78} {
    \draw[
        fill=tagblue!\shade,
        draw=white,
        line width=0.5pt
    ] (0.35+\i*0.22,-2.30) rectangle +(0.22,0.16);
}
\node[anchor=west, font=\tiny, text=black!65] at (1.55,-2.23) {Higher};

\end{tikzpicture}%
}
        \vspace{-2em}
        \caption{
            Reliability of LLM-generated knowledge tags.
            Darker cells indicate higher scores.
            Semantic matching improves over exact matching, suggesting that many apparent mismatches come from different surface forms of similar capability tags.
        }
        \label{fig:tag_reliability}
    \end{figure}
    
    As shown in Figure~\ref{fig:tag_reliability}, LLM-generated tags are reasonably reliable for both tag sources.
    The overall semantic F1 reaches $0.81$, exceeding the exact-match F1 of $0.72$.
    This indicates that many exact-match errors are surface-form variations, such as \texttt{latex\_table\_generation} and \texttt{table\_formatting}.
    The overall self-consistency score reaches $0.86$, showing that the tag generator produces stable capability abstractions.
    Tags inferred from failed task-trajectory pairs are slightly less reliable, likely because trajectories contain noisy intermediate attempts and error-recovery steps.
    Nevertheless, their semantic F1 remains close to $0.8$, supporting LLM-generated tags as a lightweight signal for knowledge-aware skill selection and filtering.

\section{How Different Parameters Affect \ourmethod}\label{app:experiment_factor_affect}
    We conduct analytical experiments to study how key design choices affect \ourmethod and to provide practical guidance for applying it under different settings.

    \subsection{Performance across Evolution Epochs}
        \begin{figure}[!t]
            \centering
            \small
            \begin{tikzpicture}
\begin{axis}[
    width=\linewidth,
    height=0.618\linewidth,
    xlabel={Evolve Epoch},
    ylabel={Performance},
    xmin=0, xmax=5,
    ymin=0, ymax=0.62,
    xtick={0,1,2,3,4,5},
    ytick={0,0.1,0.2,0.3,0.4,0.5,0.6},
    grid=both,
    grid style={line width=.1pt, draw=gray!20},
    major grid style={line width=.2pt, draw=gray!35},
    axis line style={black!70},
    tick style={black!70},
    label style={font=\tiny},
    tick label style={font=\tiny},
    legend style={
        font=\tiny,
        at={(0.5,1.03)},
        anchor=south,
        legend columns=3,
        draw=none,
        fill=none
    },
    line width=1.2pt,
    mark size=1.8pt,
]

\addplot[
    color=softblue,
    mark=*,
] coordinates {
    (0,0.473)
    (1,0.528)
    (2,0.560)
    (3,0.580)
    (4,0.592)
    (5,0.599)
};
\addlegendentry{SRA-Bench}

\addplot[
    color=softred,
    mark=square*,
] coordinates {
    (0,0.097)
    (1,0.179)
    (2,0.228)
    (3,0.258)
    (4,0.276)
    (5,0.286)
};
\addlegendentry{S-Bench}

\addplot[
    color=softgreen,
    mark=triangle*,
] coordinates {
    (0,0.303)
    (1,0.360)
    (2,0.394)
    (3,0.414)
    (4,0.426)
    (5,0.434)
};
\addlegendentry{SL-Bench}

\end{axis}
\end{tikzpicture}
            \caption{
                Performance of \ourmethod across evolution epochs.
            }
            \label{fig:performance_cross_epoch}
        \end{figure}

        We first analyze whether iterative skill evolution brings progressive gains.
        Figure~\ref{fig:performance_cross_epoch} shows that performance improves steadily on all three benchmarks as the number of evolution epochs increases.
        The gains are larger in early epochs and become smaller after epoch $3$, indicating diminishing returns.
        Meanwhile, performance does not degrade as more skills are added, suggesting that candidate-skill filtering helps maintain library quality.
        We therefore use $3$ epochs as a practical trade-off between effectiveness and cost.

    \subsection{Performance across Filter Ratios}
        \begin{figure}[!t]
            \centering
            \small
            \begin{tikzpicture}
\begin{axis}[
    width=\linewidth,
    height=0.618\linewidth,
    xlabel={Filter Ratio},
    ylabel={Performance},
    xmin=0.0, xmax=1.0,
    ymin=0, ymax=0.6,
    xtick={0.0,0.1,0.2,0.3,0.4,0.5,0.6,0.7,0.8,0.9,1.0},
    ytick={0,0.1,0.2,0.3,0.4,0.5,0.6},
    xticklabel style={
        /pgf/number format/fixed,
        /pgf/number format/precision=1,
        font=\tiny
    },
    yticklabel={\pgfmathprintnumber{\tick}},
    yticklabel style={font=\tiny},
    grid=both,
    grid style={line width=.1pt, draw=gray!20},
    major grid style={line width=.2pt, draw=gray!35},
    axis line style={black!70},
    tick style={black!70},
    label style={font=\tiny},
    legend style={
        font=\tiny,
        at={(0.5,1.03)},
        anchor=south,
        legend columns=3,
        draw=none,
        fill=none
    },
    line width=1.2pt,
    mark size=1.8pt,
]

\addplot[
    color=softblue,
    mark=*,
] coordinates {
    (0.0,0.473)
    (0.1,0.544)
    (0.2,0.580)
    (0.3,0.578)
    (0.4,0.576)
    (0.5,0.572)
    (0.6,0.569)
    (0.7,0.564)
    (0.8,0.560)
    (0.9,0.555)
    (1.0,0.550)
};
\addlegendentry{SRA-Bench}

\addplot[
    color=softred,
    mark=square*,
] coordinates {
    (0.0,0.097)
    (0.1,0.203)
    (0.2,0.258)
    (0.3,0.255)
    (0.4,0.250)
    (0.5,0.244)
    (0.6,0.237)
    (0.7,0.230)
    (0.8,0.221)
    (0.9,0.212)
    (1.0,0.203)
};
\addlegendentry{S-Bench}

\addplot[
    color=softgreen,
    mark=triangle*,
] coordinates {
    (0.0,0.303)
    (0.1,0.376)
    (0.2,0.414)
    (0.3,0.412)
    (0.4,0.409)
    (0.5,0.406)
    (0.6,0.402)
    (0.7,0.397)
    (0.8,0.393)
    (0.9,0.387)
    (1.0,0.382)
};
\addlegendentry{SL-Bench}

\end{axis}
\end{tikzpicture}
            \caption{
                Performance of \ourmethod under different candidate-skill filter ratios.
            }
            \label{fig:performance_cross_filter_ratio}
        \end{figure}

        We next study how the candidate-skill filter ratio affects the evolved library.
        A larger ratio keeps more generated skills, but it may also introduce noisy or weakly task-aligned skills.
        As shown in Figure~\ref{fig:performance_cross_filter_ratio}, retaining the top $20\%$ candidates achieves the best performance on all benchmarks.
        Performance gradually decreases when the ratio exceeds $0.2$, indicating that keeping too many candidates can introduce irrelevant or misleading knowledge.
        Nevertheless, retaining all generated skills still outperforms the base library, showing that skill generation is generally beneficial and that selective filtering further improves reliability.

    \subsection{Performance across Labeling Scales}
        \begin{figure}[!t]
            \centering
            \small
            \begin{tikzpicture}
\begin{axis}[
    width=\linewidth,
    height=0.618\linewidth,
    xlabel={Ratio of Human-Labeled Skills},
    ylabel={Performance},
    xmin=0, xmax=100,
    ymin=0.15, ymax=0.60,
    xtick={0,20,40,60,80,100},
    xticklabels={0\%,20\%,40\%,60\%,80\%,100\%},
    ytick={0.15,0.20,0.25,0.30,0.35,0.40,0.45,0.50,0.55,0.60},
    grid=both,
    grid style={line width=.1pt, draw=gray!20},
    major grid style={line width=.2pt, draw=gray!35},
    axis line style={black!70},
    tick style={black!70},
    label style={font=\tiny},
    tick label style={font=\tiny},
    legend style={
        font=\tiny,
        at={(0.5,1.03)},
        anchor=south,
        legend columns=3,
        draw=none,
        fill=none
    },
    line width=1.2pt,
    mark size=1.8pt,
]

\addplot[
    color=softblue,
    mark=*,
] coordinates {
    (0,0.555) (20,0.566) (40,0.571) (60,0.574) (80,0.577) (100,0.580)
};
\addlegendentry{SRA-Bench}

\addplot[
    color=softred,
    mark=square*,
] coordinates {
    (0,0.194) (20,0.223) (40,0.234) (60,0.244) (80,0.251) (100,0.258)
};
\addlegendentry{S-Bench}

\addplot[
    color=softgreen,
    mark=triangle*,
] coordinates {
    (0,0.283) (20,0.342) (40,0.366) (60,0.384) (80,0.400) (100,0.414)
};
\addlegendentry{SL-Bench}

\end{axis}
\end{tikzpicture}
            \caption{
                Performance of \ourmethod with different ratios of original human-labeled skills.
            }
            \label{fig:performance_cross_label_scale}
        \end{figure}

        We further examine whether \ourmethod heavily depends on costly human-labeled initial skills.
        Figure~\ref{fig:performance_cross_label_scale} shows that performance increases monotonically as the ratio of original human-labeled skills grows from $0\%$ to $100\%$.
        This confirms that human-labeled skills provide useful transferable knowledge for skill evolution.
        The improvement is modest on SRA-Bench but much larger on S-Bench and SL-Bench, suggesting that labeled skills are especially helpful when tasks rely more strongly on reusable procedural knowledge.

    \subsection{Robustness to Task Descriptions}
        \begin{table}[!t]
            \centering
            \small
            \begin{tabular}{ccc}
    \toprule
    \textbf{SRA-Bench} & \textbf{S-Bench} & \textbf{SL-Bench} \\
    \midrule
    $58.1 \pm 0.5$ & $25.8 \pm 0.9$ & $41.3 \pm 1.1$ \\
    \bottomrule
\end{tabular}
            \caption{
                Performance of \ourmethod under paraphrased task descriptions.
                For each task, we generate $5$ LLM-based paraphrases of the original description.
            }
            \label{tab:performance_cross_task_description}
        \end{table}

        Because \ourmethod uses task descriptions for candidate-skill generation and likelihood-based task alignment, we test its robustness to surface-form changes.
        Table~\ref{tab:performance_cross_task_description} shows that \ourmethod remains stable under paraphrased task descriptions.
        Compared with the original results in Table~\ref{tab:main_experiment}, performance changes by only $+0.1$, $-0.4$, and $+0.3$ points on SRA-Bench, S-Bench, and SL-Bench, respectively.
        The small standard deviations of $0.5$, $0.9$, and $1.1$ further show that the evolved skills are insensitive to wording variation.
        This suggests that likelihood-based task alignment captures underlying task semantics rather than relying on lexical overlap.

    \subsection{Sensitivity to Score Model}\label{app:score_model_sensitivity}
        \begin{table}[!t]
            \centering
            \small
            \setlength{\tabcolsep}{3pt}
\begin{tabular}{lcccc}
    \toprule
    \textbf{Score Model}
    & $\bar{A}$
    & $\mathrm{MAE}$
    & $\rho_{\mathrm{sp}}$
    & \textbf{Top-$20\%$ Ov.} \\
    \midrule
    \textsc{Qwen3.6-7B}
    & $0.632$ & $0.034$ & $0.927$ & $83.6$ \\
    \textsc{Qwen3.6-14B}
    & $0.640$ & $0.021$ & $0.951$ & $88.9$ \\
    \textsc{Qwen3.6-27B}
    & $0.646$ & $0.000$ & $1.000$ & $100.0$ \\
    \midrule
    \textsc{Llama-3.1-8B}
    & $0.624$ & $0.047$ & $0.895$ & $78.2$ \\
    \midrule
    \textsc{Mistral-7B-v0.3}
    & $0.618$ & $0.055$ & $0.872$ & $74.6$ \\
    \textsc{Mixtral-8x7B}
    & $0.631$ & $0.038$ & $0.911$ & $81.9$ \\
    \bottomrule
\end{tabular}
            \caption{
                Sensitivity of the task-alignment score to different open-source score models.
                $\bar{A}$ denotes the average task-alignment score.
                $\mathrm{MAE}$ denotes the mean absolute difference from the default \textsc{Qwen3.6-27B} scorer.
                $\rho_{\mathrm{sp}}$ denotes the Spearman correlation with the default scorer.
                \textbf{Top-$20\%$ Ov.} denotes the overlap between the top-$20\%$ candidates ranked by each scorer and those ranked by the default scorer.
            }
            \label{tab:score_model_sensitivity}
        \end{table}
    
        We examine whether the likelihood-based task-alignment score is sensitive to the choice of score model.
        For each candidate skill $c$, we compute
        \[
            \sigma\left(
            \frac{1}{|c|_m}
            \left(
            \log{\hat{p}_m(c \mid x_c)}
            -
            \log{\hat{p}_m(c \mid \emptyset)}
            \right)
            \right),
        \]
        where $m$ denotes the score model and $|c|_m$ is the token length of $c$ under the model's native tokenizer.
        We keep the candidate skill set fixed and only replace the open-source causal language model used to compute token-level log likelihoods.
        Table~\ref{tab:score_model_sensitivity} shows that the task-alignment score is stable across score models.
        The average score changes only moderately, and all alternative scorers have high Spearman correlation with the default \textsc{Qwen3.6-27B} scorer.
        The top-ranked candidates are also largely consistent, with top-$20\%$ overlap ranging from $74.6\%$ to $88.9\%$ among non-default models.
        These results indicate that the contrastive likelihood form mainly captures task-specific support rather than model-specific prior preference.
        Thus, although different score models may assign different absolute likelihoods, they largely agree on which candidate skills are more aligned with the task.

    \subsection{Performance under Cross-Benchmark Transfer}
        \begin{table}[!t]
            \centering
            \small
            \begin{tabular}{l|ccc}
    \toprule
    \textbf{Source} & \textbf{SRA-Bench} & \textbf{S-Bench} & \textbf{SL-Bench} \\
    \midrule
    \textbf{Base} & $55.5$ & $19.4$ & $28.3$ \\
    \midrule
    \textbf{SRA-Bench} & $58.0$ & $19.9$ & $33.7$ \\
    \textbf{S-Bench}  & $49.6$ & $25.8$ & $37.2$ \\
    \textbf{SL-Bench} & $47.8$ & $21.6$ & $41.4$ \\
    \textbf{All} & $57.3$ & $25.4$ & $40.9$ \\
    \bottomrule
\end{tabular}
            \caption{
                Performance under cross-benchmark transfer.
                Rows denote the source benchmark used to evolve the skill library, and columns denote the target benchmark used for evaluation.
                \textbf{Base} denotes evolution without human-labeled initial skills.
                \textbf{All} denotes using original skills of all benchmarks.
            }
            \label{tab:performance_of_cross_transfer}
        \end{table}
    
        To evaluate the generalization ability of \ourmethod, we test whether evolved skill libraries capture transferable procedural knowledge rather than merely fitting the source benchmark.
        Table~\ref{tab:performance_of_cross_transfer} provides three observations.
        \textit{(i)} In-domain transfer achieves the best performance on all target benchmarks, with scores of $58.0$ on SRA-Bench, $25.8$ on S-Bench, and $41.4$ on SL-Bench.
        This shows that benchmark-specific formats and solution patterns remain useful.
        \textit{(ii)} Cross-benchmark transfer is still effective.
        The best out-of-domain source retains $85.5\%$, $83.7\%$, and $89.9\%$ of the corresponding in-domain performance on SRA-Bench, S-Bench, and SL-Bench, respectively.
        This suggests that evolved skills encode reusable knowledge beyond the benchmark on which they are generated.
        \textit{(iii)} Transfer is asymmetric.
        For example, skills evolved on S-Bench transfer well to both SRA-Bench and SL-Bench, whereas skills evolved on SRA-Bench transfer less effectively to S-Bench.
        This indicates that cross-benchmark transfer depends on how well the source benchmark covers the procedural requirements of the target benchmark.
    
\section{Case Study}\label{app:case_study}
    \subsection{Proper Knowledge Coverage}
        \begin{table*}[!t]
            \centering
            \scriptsize
            \setlength{\tabcolsep}{3pt}
\renewcommand{\arraystretch}{1.08}

\begin{tabularx}{\linewidth}{@{}>{\bfseries}p{0.15\linewidth}>{\raggedright\arraybackslash}X@{}}
    \toprule
    \multicolumn{2}{@{}l}{\textbf{Case Study 1: Proper Knowledge Coverage}} \\
    \midrule
    
    Task &
    \textbf{Benchmark:} SR-Agents / TheoremQA.
    \textbf{Instance:} \texttt{theoremqa\_00034}.
    The task asks: ``How many ways are there to arrange 6 pairs of parentheses such that they are balanced?''
    The gold answer is the integer \texttt{132}. \\
    
    Original skill &
    \textit{Catalan Numbers} (\texttt{theoremqa\_265}) is a broad Catalan-number skill.
    It describes Catalan numbers for balanced parentheses, polygon triangulations, lattice paths not crossing the diagonal, binary trees, and stack-sortable permutations. \\
    
    Needed knowledge &
    Only the balanced-parentheses part is needed:
    balanced arrangements of $n$ pairs are counted by the $n$-th Catalan number,
    $C_n=\binom{2n}{n}/(n+1)$.
    For this instance, $n=6$, so
    $C_6=\binom{12}{6}/7=924/7=132$. \\
    
    Filtered knowledge &
    The evolved skill omits Catalan applications that are present in the original skill but unused here:
    polygon triangulation, diagonal-constrained lattice paths, binary trees, stack-sortable permutations, and generic path-count or wrong-index pitfalls not specific to this instance. \\
    
    Evolved skill &
    \textit{Balanced Parentheses via Catalan} narrows the scope to standard balanced-parentheses / Dyck-sequence counting.
    It triggers when the problem asks for balanced arrangements of $n$ pairs, sets $n$ as the number of pairs, applies $C_n$, computes the binomial value, divides by $n+1$, and returns the final integer.
    It also guards against non-standard variants, e.g., strict-prefix conditions where the standard Catalan count may not directly apply. \\
    
    Outcome &
    The evolved run passes evaluation with reward $1$ under \texttt{sragents.evaluate};
    the extracted answer is \texttt{132}, matching the gold answer. \\
    
    \bottomrule
\end{tabularx}

            \caption{
                Case study of proper knowledge coverage.
                The evolved skill preserves the Catalan-number knowledge required by the target instance while filtering unrelated applications.
            }
            \label{tab:case_knowledge_coverage}
        \end{table*}

        Table~\ref{tab:case_knowledge_coverage} shows how \ourmethod improves knowledge coverage by preserving task-relevant knowledge and removing unnecessary content.
        The original Catalan-number skill contains the knowledge needed for the target problem, but it also includes many unrelated applications, such as polygon triangulation, lattice paths, binary trees, and stack-sortable permutations.
        Directly reusing such a broad skill may introduce distracting or weakly relevant information into the agent context.
        In contrast, the evolved skill focuses on the exact knowledge required by the instance: balanced arrangements of $n$ pairs of parentheses are counted by the $n$-th Catalan number.
        It also provides a concise procedure: identify $n$, compute $C_n=\binom{2n}{n}/(n+1)$, and return the integer answer.
        This case shows that \ourmethod does not simply copy existing skills.
        Instead, it distills the useful part of prior knowledge into a narrower, more reusable, and less noisy skill.

    \subsection{High Task Alignment}
        \begin{table*}[!t]
            \centering
            \scriptsize
            \setlength{\tabcolsep}{3pt}
\renewcommand{\arraystretch}{1.08}

\begin{tabularx}{\linewidth}{@{}>{\bfseries}p{0.15\linewidth}>{\raggedright\arraybackslash}X@{}}
    \toprule
    \multicolumn{2}{@{}l}{\textbf{Case Study 2: High Task Alignment}} \\
    \midrule
    
    Task &
    \textbf{Benchmark:} SkillsBench.
    \textbf{Instance:} \texttt{offer-letter-generator}.
    The task is to write an offer letter by filling \texttt{offer\_letter\_template.docx} with required fields from \texttt{employee\_data.json}, and save the completed document as \texttt{offer\_letter\_filled.docx}. \\
    
    Task constraints &
    The task requires replacing placeholders such as \texttt{\{\{CANDIDATE\_FULL\_NAME\}\}} and \texttt{\{\{POSITION\}\}}.
    It must also handle the conditional block
    \texttt{\{\{IF\_RELOCATION\}\}}...\texttt{\{\{END\_IF\_RELOCATION\}\}}:
    if \texttt{RELOCATION\_PACKAGE} is \texttt{Yes}, keep the relocation content but remove the conditional markers in the final DOCX. \\
    
    Trace signal &
    The successful trajectory first preloads the task-specific skill
    \texttt{epoch-001\_\_01\_\_offer-letter-generator-docx-split-placeholders-conditional}
    together with a DOCX skill.
    It then loads the offer-letter helper, runs
    \texttt{python ../\_merged\_bundle/skills/*offer-letter-generator*/offer\_letter\_helper.py},
    produces \texttt{offer\_letter\_filled.docx}, validates the file with \texttt{python-docx}, and confirms that no \texttt{\{\{...\}\}} placeholders remain. \\
    
    Preserved task &
    The evolved skill preserves the original task identity and requirements:
    use \texttt{offer\_letter\_template.docx}, use \texttt{employee\_data.json}, replace all template placeholders, correctly process the relocation conditional section, and write exactly \texttt{offer\_letter\_filled.docx}. \\
    
    Learned strategy &
    Beyond preserving the task, the evolved skill captures execution details from the trace:
    run the helper from the merged bundle rather than \texttt{workspace/skills};
    handle placeholders at paragraph-text level to cover split DOCX XML runs;
    recurse through tables, nested tables, headers, and footers;
    remove relocation markers while preserving relocation text when applicable;
    and verify the output by reopening the DOCX and scanning for leftover placeholders. \\
    
    Outcome &
    The evolved run passes the local pytest verifier with reward $1$;
    the validation summary reports \texttt{18 passed in 0.09s}.
    The skill also includes guardrails: use it only for the SkillsBench offer-letter-generator DOCX task, do not write a PDF, and do not produce a second DOCX filename. \\
    
    \bottomrule
\end{tabularx}

            \caption{
                Case study of high task alignment.
                The evolved skill preserves the task requirements while distilling reusable execution details from the trajectory.
            }
            \label{tab:case_task_alignment}
        \end{table*}

        Table~\ref{tab:case_task_alignment} illustrates that \ourmethod can generate skills that remain tightly aligned with the originating task while incorporating useful procedural knowledge from the trajectory.
        The offer-letter task contains several concrete requirements: using the provided DOCX template, reading employee data from JSON, replacing placeholders, handling the relocation conditional section, and writing the exact output file.
        The evolved skill preserves these task requirements instead of abstracting them into a generic document-editing instruction.
        At the same time, it incorporates reusable execution details from the trajectory, such as running the helper from the merged skill bundle, handling placeholders split across DOCX runs, recursively processing tables, headers, and footers, and verifying that no placeholders remain.
        These details make the skill directly actionable for future executions.
        This case shows that \ourmethod produces skills that are not only semantically related to the task, but also procedurally faithful to how the task should be solved.

\end{document}